\documentclass[sigconf]{acmart}
\usepackage{multirow}
\usepackage{multirow}
\usepackage[normalem]{ulem}
\useunder{\uline}{\ul}{}
\usepackage{subcaption}
\usepackage{colortbl}

\begin{document}

\title{Generalizable Prompt Tuning for Vision-Language Models}


\author{Qian Zhang}
\affiliation{%
  \institution{Northwestern Polytechnical University}
  \city{Xi'an}
  \country{China}
}


\begin{abstract}
Prompt tuning for vision-language models such as CLIP involves optimizing the text prompts used to generate image-text pairs for specific downstream tasks. 
While hand-crafted or template-based prompts are generally applicable to a wider range of unseen classes, they tend to perform poorly in downstream tasks (i.e., seen classes). 
Learnable soft prompts, on the other hand, often perform well in downstream tasks but lack generalizability. 
Additionally, prior research has predominantly concentrated on the textual modality, with very few studies attempting to explore the prompt's generalization potential from the visual modality.
Keeping these limitations in mind, we investigate how to prompt tuning to obtain both a competitive downstream performance and generalization. 
The study shows that by treating soft and hand-crafted prompts as dual views of the textual modality, and maximizing their mutual information, we can better ensemble task-specific and general semantic information. 
Moreover, to generate more expressive prompts, the study introduces a class-wise augmentation from the visual modality, resulting in significant robustness to a wider range of unseen classes.
Extensive evaluations on several benchmarks report that the proposed approach achieves competitive results in terms of both task-specific performance and general abilities.
\end{abstract}


\maketitle

\section{Introduction}
Over the last decade, deep learning-based models for visual recognition, like VGG \cite{vgg}, ResNet \cite{resnet}, and ViT \cite{vit}, have made significant progress. 
These models are typically trained on a large dataset of image and discrete label pairs, where the label is a simple scalar generated by converting a detailed textual description to reduce the computational burden of the loss function. 
However, this paradigm has two major limitations: (1) the rich semantic relations between textual descriptions are not fully utilized, and (2) the models are limited to closed-set classes only. 
In recent years, research on large-scale Vision-Language Models (VLMs), such as Contrastive Language-Image Pretraining (CLIP) \cite{clip}, Flamingo \cite{flam}, and ALIGN \cite{align}, have shown remarkable performance in zero-shot image recognition, indicating potentials for learning open-world visual concepts with this paradigm.

\begin{table}[]
\begin{tabular}{l|ccc|ccc}
\hline
\multicolumn{1}{c|}{\multirow{2}{*}{Method}} & \multicolumn{3}{c|}{K=4}                & \multicolumn{3}{c}{K=16}                         \\ \cline{2-7} 
    \multicolumn{1}{c|}{}                        & Seen           & Unseen         & H              & Seen           & Unseen         & H              \\ \hline
\textbf{CLIP}\cite{clip}                               & \cellcolor{lightgray}65.13          & 69.02          &  66.90          & \cellcolor{lightgray}65.13          & 69.02          &  66.90          \\ \hline
\textbf{CoOp}\cite{coop}                                & 72.06          & \cellcolor{lightgray}59.69          &  65.29          & 77.24          & \cellcolor{lightgray}57.40          &  65.86          \\ \hline
\textbf{Ours}                                & \textbf{74.77} & \textbf{69.14} & \textbf{71.65} & \textbf{77.60}    & \textbf{68.36} & \textbf{72.12} \\ \hline
\end{tabular}
\caption{
CLIP with template-based prompts is applicable to a wider range of Unseen classes, but it performs poorly in downstream tasks.
CLIP with soft prompts (CoOp) performs well in downstream tasks but lack generalizability.
K denotes the number of shots per classes.
We use the abbreviation ``H'' to refer to the Harmonic Mean of ``Seen'' and ``Unseen'', which serves as a measure of efficacy.
}
\label{table1}
\end{table}

In CLIP, the model is trained to associate images and corresponding texts, such that the representations of the two modalities are close in the joint embedding space \cite{clip}. 
The pre-training process involves generating many such image-text pairs from a large corpus of images and texts.
Prompt tuning involves finding the best text prompts to use for generating these pairs. 
This can involve experimenting with different prompts to see which ones lead to the best performance on specific tasks, such as image classification or captioning \cite{prompt}. Specifically, prompt tuning can be divided into two categories:

\textbf{Hand-crafted or template-based prompts}. 
One common approach to prompt tuning is to use human evaluation or templates to adjust the prompts \cite{menon2023visual,clip}. 
For example, if the task is to recognize objects in images, prompt tuning could involve generating image-text pairs with prompts that emphasize the object category, such as ``\texttt{a photo of a [CLASS]}''. 
\texttt{[CLASS]} refers to a label like ``car''.

\textbf{Soft prompts}.
Another approach is to use algorithms or other search techniques to explore the space of possible prompts and find the ones that lead to the best performance \cite{pt1,vpt,coop,cocoop,prograd,kgcoop}.
In this approach, soft prompts are often utilized, represented by the format ``$\{$v$_1$, v$_2$, $\dots$, v$_M$, \texttt{[CLASS]}$\}$'', where v$_M$ denotes the learnable vector and is optimized by downstream tasks.

Although previous researches offer some of the advantages mentioned earlier, it also has several significant drawbacks:

1. \textbf{Trade-off}. 
While hand-crafted or template-based prompts can be applied more broadly to unseen classes, they often result in poor performance in downstream tasks (i.e., seen classes). 
Conversely, soft prompts tend to perform well in downstream tasks, but their lack of the generalizability stems from overfitting to the prompting seen classes (as referred to Table \ref{table1}).

2. \textbf{Expressiveness}. 
The expressiveness of the prompt subspace is crucial to performance and generalization.
Despite having a prompt subspace that is defined on the prompting classes, it may still be insufficient to generalize to a broader range of unseen classes.

3. \textbf{Multimodality}.
Prior research has predominantly focused on the textual modality, with very few studies attempting to explore the prompt's generalization potential from the visual modality.

\textbf{Our contributions}.
Taking these key challenges into consideration, we conduct a study on how to learn prompts to achieve both a competitive downstream performance and generalization.
We propose, for the first time, a prompt tuning method that not only ensembles task-specific and general semantic information from the textual modality but also explores the class-wise augmentation from the visual modality.
The study shows that by treating soft and hand-crafted prompts as dual views of the textual modality, and maximizing their mutual information, we can better ensemble task-specific and general semantic information. 
Moreover, given that the prompting classes are not sufficiently rich to have an expressive prompt subspace, the study introduces a class-wise augmentation from the visual modality, resulting in significant robustness to a wider range of unseen classes.
Through extensive evaluations on base-to-new generalization, domain generalization, and cross-dataset transferability settings, we show that the proposed approach achieves competitive results in terms of both task-specific and general abilities.

\section{Related Work}

\subsection{Vision-Language Models}
Recent research has demonstrated the the necessity of multimodal learning \cite{ecnet, osnet, tidnet, utdnet, c2kd, omb} that vision-language models (VLMs) trained on image-text pairs are more capable than those that rely solely on images. 
This model has received benefits from the advancements in three specific areas. 
Firstly, text representation learning using Transformers \cite{transformer} provides better feature representation abilities. 
Secondly, large-minibatch contrastive representation learning methods \cite{simclr,mom} help the network to learn discriminative information. 
Finally, web-scale training datasets \cite{clip,align}, which contain billions of image-text pairs for pretraining (0.4 billion pairs for CLIP \cite{clip}, 1.8 billion pairs for ALIGN \cite{align}), have been utilized. 
To accurately annotate image-text pairs at the billion level, unsupervised or weakly supervised learning methods \cite{simvlm} are employed in the raw training dataset. 
Specifically, ViLT \cite{vilt} improves the robustness of visual and text embedding by randomly masking words in the text. 
Additionally, masked autoencoders \cite{mae} randomly mask patches of the input image to formulate a capable self-supervised learner. 
CLIP \cite{clip} is the representative VLM, which trains the visual encoder and visual encoder using the contrastive loss based on 400 million image-text pairs obtained from the internet, exhibiting impressive zero-shot ability. 
Recent studies are devoted to improve the generalization ability of VLMs, hallucination issue \cite{sid} and OOD \cite{cocoop}.
Our study is based on adapting pre-trained VLMs to downstream applications using CLIP, following prompt tuning methods \cite{coop,cocoop,proda,kgcoop,prograd,dft}.

\subsection{Prompt Tuning}
With the development of deep learning \cite{st, ocl, tev, aipnet, evb, reqa, procc, ml2p}, pre-trained VLMs have demonstrated success in various tasks, such as point cloud understanding \cite{pointclip}, dense prediction \cite{denseclip}, continual learning \cite{continualclip1,continualclip2}, and visual understanding \cite{videoclip3}, among others. 
Prompt tuning \cite{prompt} is one of the most promising approaches to adapting pretrained VLMs to specific tasks. 
Originally introduced in Large Language Models (LLMs), prompt tuning involves hand-crafted instructions (or examples) for the task input to guide the LLMs to generate the appropriate output \cite{lmfsl}. 
Jiang et al. \cite{softprompt} replaced the hand-crafted prompts with candidate prompts obtained through text mining and paraphrasing, and during the training process, they selected the optimal candidates. 
The gradient-based approach \cite{autoprompt} selected the best tokens from the vocabulary that resulted in the most significant changes in gradients, based on the label likelihood.
In VLMs, task-related tokens are applied to infer task-specific textual knowledge through prompt tuning. 
For example, CLIP \cite{clip} uses a hand-crafted template, such as ``\texttt{a photo of a [CLASS]}'', to model the text embedding for zero-shot prediction.

However, hand-crafted prompts do not consider task-specific knowledge. 
To address this issue, CoOp \cite{coop} replaced hand-crafted prompts with learnable soft prompts optimized by few-shot labeled samples. 
But, this led to overfitting of the learnable prompts on trained tasks, which undermined their general (zero-shot) ability. 
Conditional CoOp (CoCoOp) \cite{cocoop} proposed generating conditional context vectors via an extra network and adding them to learnable prompts to alleviate the class drift problem. 
Similarly, DenseCLIP \cite{denseclip} introduced context-aware prompts via an extra Transformer module to adapt to dense prediction problems. 
ProDA \cite{proda} proposed learning output embeddings of prompts rather than input embeddings. 
Knowledge-guided CoOp (KgCoOp) \cite{kgcoop} introduced a straightforward method to minimize the Euclidean distance between embeddings of hand-crafted and learnable prompts to consider the trade-off between task-specific and general abilities. 
Visual prompting \cite{vpt,vpt2,vpt3} has also been shown to be an effective tool for large-scale visual models. 
However, these methods are limited to the Visual Transformer \cite{vit} architecture.
Vision-language prompt tuning methods \cite{vlpt1,vlpt2} generated visual and text prompts via a few learnable parameters to ensure mutual synergy.

\textbf{Our work is distinct from others in the following ways}.
Of the methods mentioned above, our work is most closely related to CoOp \cite{coop}, CoCoOp \cite{cocoop}, ProGrad \cite{prograd}, and KgCoOp \cite{kgcoop}. 
CoOp serves as the baseline method. Based on it, CoCoOp alleviates the overfitting issue by introducing extra conditional context vectors. 
Both ProGrad and KgCoOp aim to align learnable specific knowledge (i.e., learnable prompt) with general knowledge (i.e., hand-crafted prompt). 
ProGrad employs Kullback-Leibler (KL) divergence loss and modulates gradient updating to achieve this alignment, while KgCoOp constrains the euclidean distance between embedding distances of learnable and hand-crafted prompts. 
However, these two methods use hard loss functions and may not achieve optimal results for both task-specific and general abilities.
In addition to the challenge of balancing task-specific and general abilities, there is still untapped potential in exploring the visual modality.
Our work sets itself apart from others by meticulously addressing these crucial limitations, leading to superior outcomes in terms of the trade-off between task-specific and general abilities (as referred to Section~\ref{sec:exp}).

\vspace{-1em}

\subsection{Data Augmentation}
There is a substantial amount of literature on data augmentation methods to enhance the generalization ability of neural networks. 
Recently, label mixing-based approaches, such as Mixup \cite{mixup}, Cutmix \cite{cutmix}, and Saliency-Mixup \cite{salmix}, have been shown to significantly improve the generalization of neural networks. 
These mixup strategies have been successfully employed in various visual tasks, such as continual learning \cite{cl1} and long-tailed visual recognition \cite{ltvr}. 
Despite their success, current vision-language prompt tuning methods have not focused on utilizing image augmentation techniques to enhance generalization ability. 
Interestingly, previous methods that adopted the same Mixup strategy as ours did not show significant improvements and even performed worse in some metrics, as demonstrated in Table~\ref{table2}. 
One potential reason for this outcome is that, with only learnable parameters in the text inputs, the uninformative pixels generated by Mixup \cite{mixup} could mislead existing methods and cause them to learn unexpected feature representations, even noises in the under-parameterized regime \cite{mixup_reason}. 
Possible solutions to this problem could involve advanced augmentation strategies \cite{cutmix,salmix} that selectively mix informative regions of two classes, or update the visual backbone, which could potentially disturb the pre-trained parameters and cause misalignment between visual and textual information. 
In this work, we innovatively employ a learnable mutual information estimator to maximize the mutual information of dual views of augmented features, which enables us to balance task-specific and general abilities and effectively leverage the augmented visual modality.

\subsection{Knowledge Ensemble}
Knowledge distillation and sample replay are commonly used in continual learning \cite{cl1,cl3} to mitigate forgetting and adapt to new knowledge. 
However, prompt tuning for VLMs differs from continual learning as it assumes that the VLMs have already acquired all the necessary knowledge from the pre-training phase and focus on creating a task-specific query. 
Furthermore, prompt tuning does not have access to the pre-trained data and does not require memory buffers to be revisited frequently. 
While previous methods \cite{prograd,kgcoop} for prompt tuning used hard loss function constraints to ensemble task-specific and general knowledge, these achieved sub-optimal trade-offs (as referred to Section~\ref{sec:exp}). 
Drawing inspiration from self-supervised representation learning \cite{ssl} that enforces the network to learn invariant feature representations, we treat the embeddings of learnable and hand-crafted prompts as dual views of the text modality and encourage the network to extract shared semantic information by maximizing mutual information. 
Our approach softly ensembles task-specific and general information and can also accommodate augmented features.

\section{Methodology}
\begin{figure*}[t]
\centering
\includegraphics[width=0.8\textwidth]{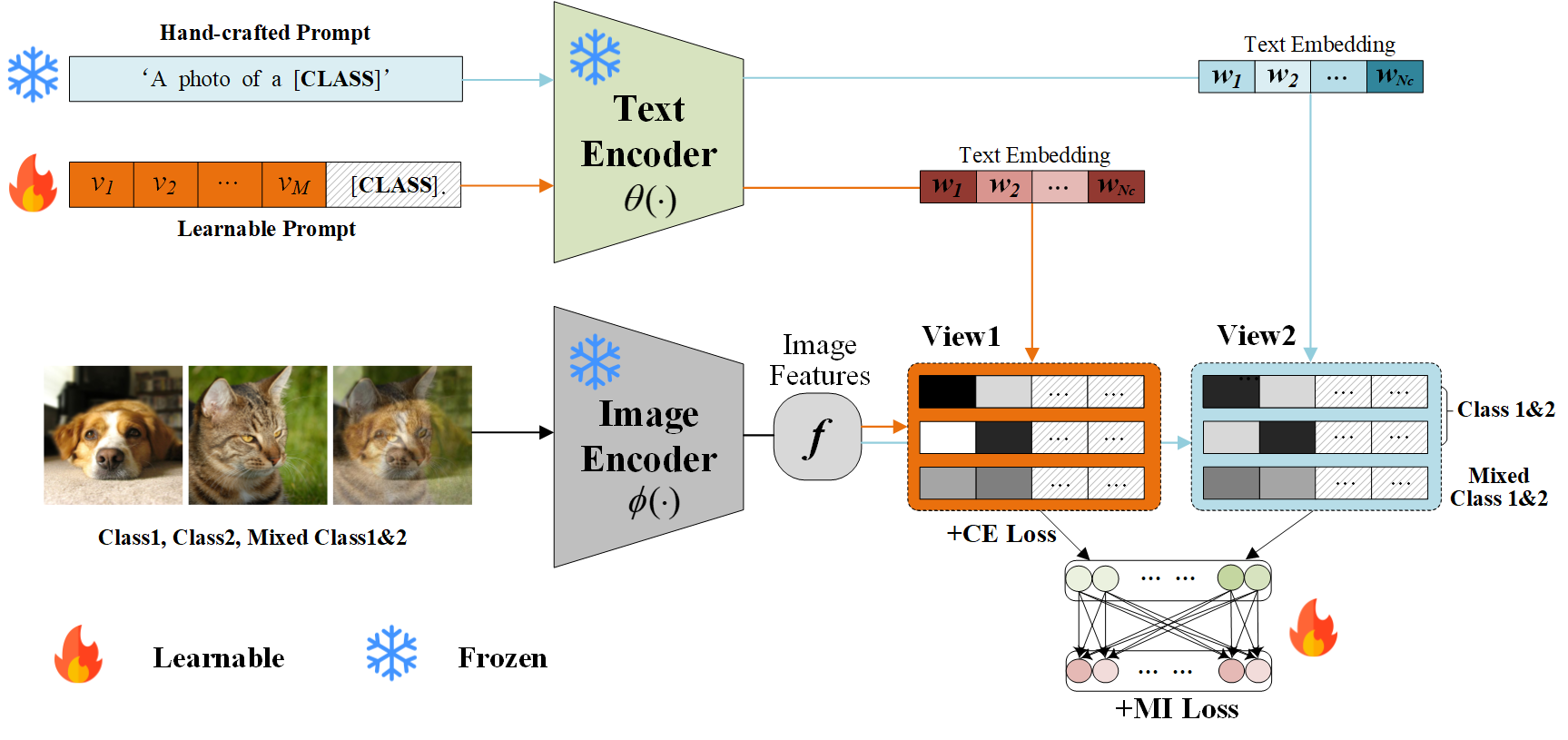}
\vspace{-1.2em}
\caption{
The framework overview. We illustrate our method that is fine-tuned on two classes for the sake of simplicity. 
We start by fusing the image features from Class 1, Class 2, and Mixed Class 1$\&$2 with the text embeddings of hand-crafted and learnable prompts, respectively. 
This process allows us to create augmented dual views of the prediction probability. 
Next, the learnable MI estimator is utilized to ensemble the shared semantic cues from both general and specific information.
}
\label{fig1}
\vspace{-0.3em}
\end{figure*}

Figure~\ref{fig1} provides an overview of the proposed approach. 
In this section, we first provide a brief review of prompt-based zero-shot inference and learnable soft prompt tuning for the CLIP \cite{clip} model. 
Next, we discuss the motivations behind our proposed method in detail. Finally, we provide a comprehensive introduction to our proposed approach.


\subsection{Preliminaries}

\textbf{Zero-shot Inference.} 
CLIP has a strong zero-shot ability \cite{clip,coop} due to being pre-trained on minimizing the distance between visual and textual embedding space.
Zero-shot inference using the pre-trained CLIP model aims to infer downstream tasks without fine-tuning the model.
For instance, in the classification task, zero-shot inference treats it as an image-text matching problem, where text inputs are obtained by a textual template \emph{\textbf{t}}$^{clip}$ based on the predicted classes.
Formally, given the pre-trained vision encoder $\phi(\cdot)$ and text encoder $\theta(\cdot)$, the text embeddings $\mathbf{W}^{clip}=\left\{\mathbf{w}_{i}^{clip}\right\}_{i=1}^{N_{c}}$ are generated, where the text embeddings of $i$-th class are denoted by $\mathbf{w}_{i}^{clip}$= $\theta$(\emph{\textbf{t}$^{clip}_{i}$}). $N_{c}$ represents the total number of classes for the downstream task.
Then, image features (\emph{\textbf{f}}) are extracted from the input image ($\boldsymbol{x}$), and the image-text matching score is measured using cosine similarity $\left\langle\mathbf{w}_{i},\emph{\textbf{f}}\right\rangle$.
The prediction probability for the $i$-th class is obtained in this way:
\begin{eqnarray}
p_{\mathrm{zs}}^{clip}\left(\boldsymbol{w}_i\mid\boldsymbol{x}\right)=\frac{\exp\left(<\textbf{w}_i, \boldsymbol{f}>/\tau\right)}{\sum_{j=1}^{N^c}\exp\left(<\textbf{w}_j, \boldsymbol{f}>/\tau\right)},
\end{eqnarray}
where $\tau$ is a temperature scalar \cite{clip}.

\textbf{Learnable Prompt Tuning.}
The use of hand-crafted prompts can reduce the distribution gap between pre-training text inputs, but they often lead to poor performance on specific downstream tasks.
To address this issue, Context Optimization (CoOp) \cite{coop} utilizes a set of continuous context vectors to generate task-specific textual embeddings in an end-to-end manner.
Specifically, CoOp incorporates $M$ context vectors and constructs the prompt as $\boldsymbol{t_i}=\{\boldsymbol{v_1, v_2, \dots, v_M, c_i}\}$, where $\boldsymbol{v}=\{\boldsymbol{v_1, v_2, \dots, v_M}\}$ represents the learnable context vectors
and $\boldsymbol{c_i}$ represents the $i$-th class token embedding.
The learnable prompt is then passed to the text encoder $\theta(\cdot)$. 
The prediction probability for the $i$-th class is obtained:
\begin{eqnarray}
\quad {p}^{coop}\left(\boldsymbol{t}_i \mid \boldsymbol{x}\right)=\frac{\exp \left(<\theta\left(\boldsymbol{t}_i\right), \boldsymbol{f}>/ \tau\right)}{\sum_{j=1}^{N^c} \exp \left(<\theta\left(\boldsymbol{t}_j\right), \boldsymbol{f}>/ \tau\right)},
\end{eqnarray}
Then, the few-shot ground-truth labels ($\boldsymbol{y}$) serve as the task-specific knowledge, and CoOp optimizes the learnable context tokens $\boldsymbol{v}$ via the cross-entropy $\mathcal{L}_{\mathrm{ce}}$ as follows:
%
\begin{eqnarray}
\mathcal{L}_{\mathrm{ce}}(\boldsymbol{v})=-{\sum_{i=1}^{N^c}} \boldsymbol{y}_i \log {p}^{coop}\left(\boldsymbol{t}_i \mid \boldsymbol{x}\right).
\end{eqnarray}

\subsection{Motivations}
While CoOp-based methods are effective in adapting pre-trained CLIP to downstream tasks, they may cause biases towards the base classes with few-shot training samples, which in turn, negatively impacts generalization ability. 
To address this issue, Prompt-aligned Gradient (Prograd) \cite{prograd} has been proposed, which utilizes zero-shot CLIP predictions as general knowledge to guide the optimization direction. 
Specifically, Prograd employs Kullback-Leibler (KL) divergence between CoOp's prediction and the zero-shot CLIP prediction:
\begin{eqnarray}
\mathcal{L}_{\mathrm{kl}}(\boldsymbol{v})=-{\sum_{i=1}^{N^c}} \boldsymbol p_{\mathrm{zs}}^{clip}\left(\boldsymbol{w}_i \mid \boldsymbol{x}\right) \log \frac{{p}^{coop}\left(\boldsymbol{t}_i \mid \boldsymbol{x}\right)}{p_{\mathrm{zs}}^{clip}\left(\boldsymbol{w}_i \mid \boldsymbol{x}\right)}.
\end{eqnarray}

One straightforward approach employed by Knowledge-guided Context Optimization (KgCoOp) \cite{kgcoop} is to minimize the Euclidean distance $\left | \left | \cdot  \right |  \right |$ between hand-crafted text embeddings and learnable text embeddings:
\begin{eqnarray}
\mathcal{L}_{kg}(\boldsymbol{v})=\frac{1}{N_{c}} \sum_{j=1}^{N_{_{c}}}  \left | \left | \boldsymbol{w}_{j}-\theta(\boldsymbol{t}_{j})   \right |  \right |_{2}^{2}.
\end{eqnarray}

Although KgCoOp and ProGrad attempted to address the overfitting problem, they did not achieve the optimal balance between general and specific information. 
Specifically, KgCoOp's use of the Euclidean distance constraint impedes the learnable prompt's ability to adapt to task-specific information, resulting in weak performance on base classes. 
For example, in the 16-shot setting, the accuracy of the base class drops by 1.73$\%$ compared to CoOp. 
On the other hand, ProGrad's use of the KL loss, which is a relatively soft regularization compared to the Euclidean distance, may not fully incorporate the general information from hand-crafted prompts. 
In the 16-shot setting, the accuracy of new classes is 3.12$\%$ lower than KgCoOp (as referred to Section~\ref{sec:exp}).

Furthermore, the existing methods neglect sufficient exploration of image inputs. 
Conditional CoOp (CoCoOp) \cite{cocoop} addresses this issue by generating a conditional vector for each image with an additional neural network to shift the focus from a specific set of classes to each input sample. 
However, this conditional design requires an independent forward pass for sample-specific prompts through the text encoder, resulting in significant computation burdens. 
Additionally, when incorporating hand-crafted general information, we believe that ``cross-class information also really matters'', encouraging the network to shift attention to more augmentation classes rather than a fixed set of prompting classes.
To overcome these two issues, we propose a novel method and provide detailed explanations below.

\subsection{Fusion between Textual Modal Ensemble and Visual Modal Exploration}

\textbf{Textual Modal Ensemble with MI Maximization}.
%
%
To address the first issue discussed earlier, we propose a different approach that treats the hand-crafted and learnable prompts as two complementary views of the text modality, and we maximize the Mutual Information (MI) between them to effectively extract shared semantic information. 
MI is a fundamental measure of the relationship between random variables \cite{ssl}, and in information theory, the integration of two variables X$_1$ and X$_2$ can be achieved by maximizing their MI, as this ensures that the most relevant information is retained:
\begin{eqnarray}
MI(X_1, X_2)=H(X_1)-H(X_1|X_2),
\end{eqnarray}
where $H(\cdot)$ and $H(\cdot|\cdot)$ represent the information entropy and the conditional entropy. 
Equivalently, $MI(X_1, X_2)$ can also be reformulated by:
\begin{eqnarray}
MI(X_1, X_2)=MI(X_2, X_1)=H(X_2)-H(X_2|X_1)\\ \nonumber
=H(X_1, X_2)-H(X_1|X_2)-H(X_2|X_1).
\end{eqnarray}
Given Equations (6) and (7), $MI(X_1, X_2)$ can be reformulated by:
\begin{eqnarray}
MI(X_1, X_2)= \frac{1}{3}\left[H(X_1) + H(X_2) + {H(X_1, X_2)}\right]\\ \nonumber
-\frac{2}{3}\left[H(X_1|X_2)+H(X_2|X_1)\right].
\end{eqnarray}
The conditional entropy $H(X_1|X_2)$ is expressed as:
\begin{eqnarray}
H(X_1|X_2)=\sum_{x_1 \in X_1, x_2 \in X_2} p\left(x_1, x_2\right) \log \frac{p\left(x_1, x_2\right)}{p\left(x_2\right)},
\end{eqnarray}
where $p(x)$ indicates the marginal distributions of $x$. 
$p(x_1, x_2)$ represents the joint distribution of $x_1$ and $x_2$. 
As $x_1$ and $x_2$ are dual views of the same set of classes, we force the joint distribution to be the same as the marginal distributions.
$MI(X_1, X_2)$ can be approximated by:
\begin{eqnarray}
MI(X_1, X_2)\approx \frac{1}{3}\left[H(X_1) + H(X_2) + H(X_1, X_2)\right].
\end{eqnarray}

To estimate the joint probability matrix $\boldsymbol{P}\in\mathbb{R}^{C\times C}$, we follow Invariant Information Clustering \cite{iic} to formulate our MI estimator. 
Given the small batch $B_n$, the matrix $\boldsymbol{P}$ can be computed by:
\begin{eqnarray}
\boldsymbol{P}=\frac{1}{n} \sum_{i=1}^n \sigma \left[\varphi \left(\boldsymbol{x}_i^1\right)\right] \cdot \sigma \left[\varphi\left(\boldsymbol{x}_i^2\right)\right]^{\top},
\end{eqnarray}
where $\boldsymbol{x}_i^1$ and $\boldsymbol{x}_i^2$ are the hand-crafted and learnable embeddings of the same text prompt $\boldsymbol{x}_i$,
and $\sigma \left[\varphi (\boldsymbol{x}_i^1)\right]=\text{softmax}(z) \in [0,1]^C$,
which can be interpreted as the distribution of a discrete random variable $z$ over $C$ classes, i.e., $\boldsymbol{P}(z=c|x)=\sigma \left[\varphi(\boldsymbol{x})\right]$.
The element in $\boldsymbol{P}$ at the $c_1$-th row and $c_2$-th column constitutes the joint probability $\boldsymbol{P}_{c_1c_2}=P(z_1=c_1,z_2=c_2)$. 
The marginal distributions can be obtained by summing
over the rows and columns of the matrix $\boldsymbol{P}$, i.e., $\boldsymbol{P}_{c_1}=P(z_1 = c_1)$, and $\boldsymbol{P}_{c_2}=P(z_2 = c_2)$. 
As we generally consider symmetric problems and $\boldsymbol{P}_{c_1c_2} = \boldsymbol{P}_{c_2c_1}$, $\boldsymbol{P}$ is symmetrized by $\boldsymbol{P}_{c_1c_2} = \frac{\boldsymbol{P}+\boldsymbol{P}^{\top}}{2}$. 
Following \cite{iic}, we formulate a learnable two-layer Multi-Layer Perceptron (MLP) as the MI estimator to estimate the joint distribution. 
Moreover, to make the approximate calculation of MI in Equation (8) hold, we apply the distance constrains $L_{dc}$ between the joint distribution and the marginal distributions:
\begin{eqnarray}
\mathcal{L}_{dc}(z_1, z_2)=KL(p(z_1, z_2), p(z_1)) + KL(p(z_1, z_2), p(z_2)),
\end{eqnarray}
where $KL$ indicates the Kullback-Leibler (KL) divergence loss.

Formally, the total loss function $\mathcal{L}$ is formulated as:
\begin{eqnarray}
\mathcal{L} = \mathcal{L}_{ce}(\boldsymbol{v}) + \lambda _1  \mathcal{L}_{MI}(\boldsymbol{w}, \theta (\boldsymbol{t})) + \lambda _2 \mathcal{L}_{dc}(\boldsymbol{w}, \theta (\boldsymbol{t})),
\end{eqnarray}
where $\boldsymbol{w}$ and $\theta (\boldsymbol{t})$ represent the embeddings of hand-crafted and learnable prompts, respectively. The first term in Equation (13) aims to optimize the learnable context tokens $\boldsymbol{v}$ like the CoOp-based method in Equation (2). The second and third terms are devoted to ensemble the general and specific information via Mutual Information Maximization (MIM). 

\textbf{Visual Modal Exploration with Class-wise Augmentation.} 
As mentioned previously, CoCoOp \cite{cocoop} introduced learnable sample-wise information to mitigate the overfitting issue. 
However, when incorporating hand-crafted general information, we believe that cross-class information is more important. 
This is due to two reasons: 
1) augmented classes encourage the network to pay attention to a broader range of unseen classes rather than just the fixed set of prompting classes, and 
2) as Equation (2) shows, the text prompt modulates the image embedding in a class-wise manner. 
By augmenting the classes, more general cues from hand-crafted prompts are transferred to learnable prompts through MI maximization. 
To accomplish this, we adopt Mixup \cite{mixup} for cross-class augmentation. 
Given two samples ${x_a}$ and ${x_b}$ from two classes labeled $y_a$ and $y_b$ ($a\ne b$), Mixup augments ${x_a}$ and ${x_b}$ to produce a new training sample $x_{ab}^{new}$ labeled $y_{ab}^{new}$:
\begin{eqnarray}
x_{ab}^{new}=\lambda  x_{a}+(1-\lambda ) x_{b}, \\ \nonumber
y_{ab}^{new}=\lambda  y_{a}+(1-\lambda ) y_{b}. 
\end{eqnarray}
where we restrict $\lambda$ sampling from the interval $[0.4, 0.6]$ to reduce the overlap between the augmented and original classes.

One may ask: ``\emph{What is the affecting of class-wise augmentation for CoOp-based methods and other approaches that employ hand-crafted general knowledge}?''
To answer this question, we perform experiments and report the results in Table~\ref{table2}.
As KgCoOp \cite{kgcoop} only optimizes the text embedding and does not interact with image features, we focus our analysis on CoOp \cite{coop} and ProGrad \cite{prograd}. 
We find that the Mixup strategy improves the generalization of ProGrad but negatively affects the base classes to some extent.
This is because Mixup mixes two samples without discrimination, leading the classifier to learn unexpected feature representations \cite{cutmix, salmix}, even noises in the under-parameterized regime \cite{mixup_reason}. 
ProGrad uses hand-crafted general knowledge to alleviate overfitting but fails in base classes. 
On the other hand, without hand-crafted general knowledge, CoOp fails in both base and new classes.

\begin{table}[]
\small
\caption{We compare the performance of the class-wise augmentation strategy using 4 and 16-shot, and report the average performance across all 11 datasets.}
\vspace{-1em}
\begin{tabular}{l|cll|cll}
\hline
\multicolumn{1}{c|}{}                         & \multicolumn{3}{c|}{K=4}                                                           & \multicolumn{3}{c}{K=16}                                                                              \\ \cline{2-7} 
\multicolumn{1}{c|}{\multirow{-2}{*}{Method}} & Base                      & \multicolumn{1}{c}{New}   & \multicolumn{1}{c|}{H}     & Base                 & \multicolumn{1}{c}{Novel}                         & \multicolumn{1}{c}{H}      \\ \hline
CoOp                                          & 72.06                     & \multicolumn{1}{c}{59.69} & \multicolumn{1}{c|}{65.29} & 77.24                & \multicolumn{1}{c}{57.40}                         & \multicolumn{1}{c}{65.86}  \\
+ Mixup \cite{mixup}                                   & \multicolumn{1}{l}{71.86} & 58.72                     & 64.53                      & \multicolumn{1}{l}{76.88} & 56.72                                                  &  65.06                          \\
$\bigtriangleup $                                          & \multicolumn{1}{l}{-0.20} & -0.97                     & -0.76                      & \multicolumn{1}{l}{-0.36} &  -0.68                                                 & -0.80                           \\
+ SalMix  \cite{salmix}                                & \multicolumn{1}{l}{72.35} & 61.02                     & 66.13                      & \multicolumn{1}{l}{77.52} & 58.41                                                  & 66.41                           \\
$\bigtriangleup $                                           & +0.29                     & \multicolumn{1}{c}{+1.33} & \multicolumn{1}{c|}{+0.94} & +0.28                     & \multicolumn{1}{c}{+1.01}                              & \multicolumn{1}{c}{+0.55}       \\ \hline
ProGrad                                       & 73.88                     & \multicolumn{1}{c}{64.95} & \multicolumn{1}{c|}{69.13} & 77.98                & \multicolumn{1}{c}{64.41} & \multicolumn{1}{c}{69.94}  \\
+ Mixup \cite{mixup}                               & \multicolumn{1}{l}{73.15} & 66.21                     & 68.67                      & \multicolumn{1}{l}{76.44} &   65.22                                                &  69.82                          \\
$\bigtriangleup $                                           & \multicolumn{1}{l}{-0.73} & +1.26                     & +0.46                      & \multicolumn{1}{l}{-1.54} &  +0.81                                                 & -0.12                           \\
+ SalMix \cite{salmix}                              & \multicolumn{1}{l}{74.50} & 66.91                     & 70.45                      & \multicolumn{1}{l}{77.89} & 65.62                                                 &70.55                             \\
$\bigtriangleup $                                           & +0.62                     & \multicolumn{1}{c}{+1.96} & \multicolumn{1}{c|}{+1.32} & -0.09                     & \multicolumn{1}{c}{+1.21}                        & \multicolumn{1}{c}{+0.61} \\ \hline
Ours(Mixup)                                   & 74.77                     & \multicolumn{1}{c}{69.14} & \multicolumn{1}{c|}{71.65} & 77.60                & \multicolumn{1}{c}{68.36}                         & \multicolumn{1}{c}{72.12}  \\ 
Ours(w/o Mixup)                               & 74.23                     & \multicolumn{1}{c}{68.23} & \multicolumn{1}{c|}{71.08} &  77.21              & \multicolumn{1}{c}{67.43}                    & \multicolumn{1}{c}{71.56}  \\ 
Ours(SalMix)                                   & 74.97                     & \multicolumn{1}{c}{69.81} & \multicolumn{1}{c|}{71.98} & 77.92             & \multicolumn{1}{c}{68.72}                     & \multicolumn{1}{c}{72.59}  \\ \hline
\end{tabular}
\label{table2}
\vspace{-1em}
\end{table}

To improve the performance of CoOp and ProGrad, we adopt Saliency-based Mixup (SalMix) \cite{salmix} to mix two samples based on saliency regions and aggregate more appropriate feature representations. 
We observe significant improvements in the performance of both CoOp and ProGrad, demonstrating the effectiveness of the class-wise augmentation strategy. 
Unlike CoCoOp \cite{cocoop} and DenseCLIP \cite{denseclip}, which require an extra network to induce context-aware prompt, our class-wise augmentation strategy can enrich contextual information in a cost-effective way.

Our proposed method benefits from the mixup strategy in both base and new classes since our learnable MI estimator allows augmented image features to interact effectively with task-specific and general text embeddings to learn informative features. 
Note that our MI estimator does not add much computation burden (see Section~\ref{subsec:te}) and is not used during the inference phase.

\section{Experiment} \label{sec:exp}
In accordance with CoCoOp \cite{cocoop}, we assess the efficacy of our approach in terms of three aspects: 
1. the generalization ability from the base (seen) classes to new (unseen) classes within a dataset [Section~\ref{subsec:base-new}];
2. domain generalization ability [Section~\ref{subsec:domain}]; and 
3. cross-dataset transferability [Section~\ref{subsec:cross}]. 
All our experiments utilize the pre-trained CLIP \cite{clip}.


\textbf{Datasets.} 
We evaluate our proposed approach against the first and third aspects, following the settings used in \cite{clip, coop, cocoop, prograd, kgcoop}, and conduct experiments on 11 datasets:
%
ImageNet\cite{imagenet} and Caltech101 \cite{caltech} for generic object classification; 
OxfordPets \cite{pets}, Flowers102\cite{flowers}, StanfordCars \cite{cars}, FGVCAircraft \cite{craft} , and Food101 \cite{food} for fine-grained classification; 
UCF101 \cite{ucf101} for action recognition; 
EuroSAT \cite{sat} for satellite image classification; 
SUN397 \cite{sun} for scene classification; 
DTD \cite{dtd} for texture classification. 
For the second aspect, we examine domain generalization by using ImageNet and its variants, with ImageNet serving as the source domain and its variants, including ImageNetV2 \cite{v2}, ImageNet-Sketch \cite{sketch}, ImageNet-A \cite{a}, and ImageNet-R \cite{r}, as target domains. 
Please refer to the Appendix for further information on each dataset.

\begin{table*}[t]
\caption{
Accuracy ($\%$) comparisons among all 11 datasets. 
K: the number of shots. 
``Base'' and ``New'': the classes that are ``seen'' and ``unseen'' within a dataset. 
H: Harmonic mean. 
%
%
The best two results are marked in \textbf{Bold} and \underline{Underline} (same below).
}
\vspace{-1em}
\begin{tabular}{l|ccc|ccc|ccc|ccc|ccc}
\hline
\multicolumn{1}{c|}{\multirow{2}{*}{\textbf{Methods}}} & \multicolumn{3}{c|}{\textbf{K=1}}                & \multicolumn{3}{c|}{\textbf{K=2}}                & \multicolumn{3}{c|}{\textbf{K=4}}                & \multicolumn{3}{c|}{\textbf{K=8}}                & \multicolumn{3}{c}{\textbf{K=16}}                \\ \cline{2-16} 
\multicolumn{1}{c|}{}                                  & Base           & New            & H              & Base           & New            & H              & Base           & New            & H              & Base           & New            & H              & Base           & New            & H              \\ \hline
CoOp                                                   & 58.57          & 53.51         & 55.92          & 64.75          & 52.63          & 58.07          & 72.06          & 59.69          & 65.29          & 74.72          & 58.05          & 65.34          & 77.24          & 57.40          & 65.86          \\
CoCoOp                                                 & 64.13          & 60.42          & 62.22          & 67.70          & 59.36          & 63.25          & 71.39          & 65.74          & 68.45          & 73.40          & 66.42          & 69.29          & 75.20          & 63.64          & 68.90          \\
ProGrad                                                & {\ul 68.73}    & 64.98          & {\ul 66.81}    & {\ul 71.42}    & 64.00          & 67.50          & {\ul 73.88}    & 64.95          & 69.13          & \textbf{76.25} & 64.74          & 70.03          & \textbf{77.98} & 64.41          & 69.94          \\
KgCoOp                                                 & 65.79          & {\ul 65.98}    & 65.86          & 69.67          & {\ul 67.35}    & {\ul 68.62}    & 72.42          & {\ul 68.00}    & {\ul 70.14}    & 74.08          & {\ul 67.86}    & {\ul 70.84}    & 75.51          & {\ul 67.53}    & {\ul 71.30}    \\ \hline
Ours                                                   & \textbf{70.76} & \textbf{68.43} & \textbf{68.61} & \textbf{72.06} & \textbf{69.03} & \textbf{69.70} & \textbf{74.77} & \textbf{69.14} & \textbf{71.65} & {\ul 75.90}    & \textbf{68.75} & \textbf{71.61} & {\ul 77.60}    & \textbf{68.36} & \textbf{72.12} \\ \hline
\end{tabular}
\label{table3}
\vspace{-1em}
\end{table*}

\begin{table*}[t]
    \caption{
    Per-dataset view.
    %
    Prompts are tuned with 4 shots on base classes. 
    %
    } 
    \vspace{-1em}
    \begin{subtable}{.33\linewidth}
      \centering
        \caption{Average over 11 datasets.}
        \resizebox{!}{1.3cm}{
        \begin{tabular}{l|cc|c}
            \toprule
             & Base & New & H \\
            \midrule
            CLIP & 65.13 & 69.02  & 66.90\\
            CoOp & 71.97 & 57.62  & 62.95\\
            CoCoOp & 71.81 & 60.09 &64.90 \\
            ProGrad & \underline{73.99} &63.72  &67.91 \\
            KgCoOp & 72.42 &\underline{68.00}  & \underline{70.14}\\\hline 
            Ours & \textbf{74.77} &\textbf{69.14}  &\textbf{71.65} \\
            \bottomrule
        \end{tabular}
        }
    \end{subtable}%
    \begin{subtable}{.33\linewidth}
      \centering
        \caption{ImageNet.}
        \resizebox{!}{1.3cm}{
        \begin{tabular}{l|cc|c}
            \toprule
             & Base & New & H \\
            \midrule
            CLIP & 64.46 & \underline{59.99}  & 62.14\\
            CoOp & 64.58 & 55.81 & 59.88\\
            CoCoOp & 66.28 & 58.68 &62.25 \\
            ProGrad & \underline{66.40} &58.39  &62.14 \\
            KgCoOp & 64.78 &59.68  & \underline{62.67}\\\hline 
            Ours & \textbf{66.80} &\textbf{60.90}  &\textbf{63.71} \\
            \bottomrule
        \end{tabular}
        }
    \end{subtable}%
    \begin{subtable}{.33\linewidth}
      \centering
        \caption{Caltech101.}
        \resizebox{!}{1.3cm}{
        \begin{tabular}{l|cc|c}
            \toprule
             & Base & New & H \\
            \midrule
            CLIP & 90.90 & 90.72  & 90.81\\
            CoOp & 94.06 & 87.41 & 90.61\\
            CoCoOp & \textbf{94.23} & 86.90 &90.42 \\
            ProGrad & \underline{94.10} &89.01  &91.48 \\
            KgCoOp & 93.38 &\underline{91.01}  & \underline{92.18}\\\hline 
            Ours & 93.81 &\textbf{92.38}  &\textbf{93.08} \\
            \bottomrule
        \end{tabular}
        }
    \end{subtable}%

        \begin{subtable}{.33\linewidth}
      \centering
        \caption{OxfordPets.}
        \resizebox{!}{1.3cm}{
        \begin{tabular}{l|cc|c}
            \toprule
             & Base & New & H \\
            \midrule
            CLIP & 90.11 & 94.30  & 92.16\\
            CoOp & 89.12 & 91.84 & 90.46\\
            CoCoOp & 89.46 & 90.04 &89.75 \\
            ProGrad & \underline{91.65} &\underline{95.13}  &\underline{93.36} \\
            KgCoOp & 91.50 &94.52  & 92.99\\\hline 
            Ours & \textbf{91.87} &\textbf{95.69}  &\textbf{93.74} \\
            \bottomrule
        \end{tabular}
        }
    \end{subtable}%
    \begin{subtable}{.33\linewidth}
      \centering
        \caption{StanfordCars.}
        \resizebox{!}{1.3cm}{
        \begin{tabular}{l|cc|c}
            \toprule
             & Base & New & H \\
            \midrule
            CLIP & 55.55 & \underline{66.35}  & 60.47\\
            CoOp & 61.54 & 58.62 & 60.04\\
            CoCoOp & 60.77 & 54.68 &57.56 \\
            ProGrad & \textbf{64.86} &60.77  &62.75 \\
            KgCoOp & 62.67 &\textbf{66.82}  & \underline{64.68}\\\hline 
            Ours & \underline{64.32} &64.79  &\textbf{65.53} \\
            \bottomrule
        \end{tabular}
        }
    \end{subtable}%
    \begin{subtable}{.33\linewidth}
      \centering
        \caption{Flowers102.}
        \resizebox{!}{1.3cm}{
        \begin{tabular}{l|cc|c}
            \toprule
             & Base & New & H \\
            \midrule
            CLIP & 68.47 & \textbf{73.90}  & 71.08\\
            CoOp & 88.04 & 58.23 & 70.10\\
            CoCoOp & 87.75 & 62.08 &72.72 \\
            ProGrad & \underline{89.43} &70.28  &\underline{78.71} \\
            KgCoOp & 85.42 &72.04  & 78.16\\\hline 
            Ours & \textbf{90.75} &\underline{72.27}  &\textbf{81.69} \\
            \bottomrule
        \end{tabular}
        }
    \end{subtable}%

        \begin{subtable}{.33\linewidth}
      \centering
        \caption{Food101.}
        \resizebox{!}{1.3cm}{
        \begin{tabular}{l|cc|c}
            \toprule
             & Base & New & H \\
            \midrule
            CLIP & \underline{83.71} & 84.76  & 84.23\\
            CoOp & 78.03 & 76.97 & 77.50\\
            CoCoOp & 78.78 & 77.43 &78.10 \\
            ProGrad & 81.01 &81.04  &81.02 \\
            KgCoOp & 82.82 &\textbf{85.15}  & \underline{83.97}\\\hline 
            Ours & \textbf{83.84} &\underline{84.81}  &\textbf{84.32} \\
            \bottomrule
        \end{tabular}
        }
    \end{subtable}%
    \begin{subtable}{.33\linewidth}
      \centering
        \caption{FGVCAircraft.}
        \resizebox{!}{1.3cm}{
        \begin{tabular}{l|cc|c}
            \toprule
             & Base & New & H \\
            \midrule
            CLIP & 19.27 & 26.45  & 22.30\\
            CoOp & 20.03 & 10.82 & 14.05\\
            CoCoOp & 16.69 & 17.72 &17.19 \\
            ProGrad & 23.87 &20.24  &21.91 \\
            KgCoOp & \underline{23.95} &\underline{26.86}  &\underline{25.32} \\\hline 
            Ours & \textbf{29.28} &\textbf{27.22}  &\textbf{28.21} \\
            \bottomrule
        \end{tabular}
        }
    \end{subtable}%
    \begin{subtable}{.33\linewidth}
      \centering
        \caption{EuroSAT.}
        \resizebox{!}{1.3cm}{
        \begin{tabular}{l|cc|c}
            \toprule
             & Base & New & H \\
            \midrule
            CLIP & 55.81 & \textbf{66.87}  & 60.84\\
            CoOp & 85.18 & 33.74 & 48.33\\
            CoCoOp & \underline{86.39} & 50.85 &64.02 \\
            ProGrad & \textbf{87.04} &44.67  & 59.04 \\
            KgCoOp & 79.88 &57.62  &\underline{66.75}\\\hline
            Ours & 81.29 &\underline{61.49}  &\textbf{70.02} \\
            \bottomrule
        \end{tabular}
        }
    \end{subtable}%
    
    \begin{subtable}{.33\linewidth}
      \centering
        \caption{SUN397.}
        \resizebox{!}{1.3cm}{
        \begin{tabular}{l|cc|c}
            \toprule
             & Base & New & H \\
            \midrule
            CLIP & 66.47 & 70.17  & 68.27\\
            CoOp & 71.04 & 61.06 & 65.67\\
            CoCoOp & 69.86 & 64.90 &67.29 \\
            ProGrad & \underline{73.00} &67.83  &70.32 \\
            KgCoOp & 71.82 &\underline{71.77}  & \underline{71.79}\\\hline
            Ours & \textbf{73.12} &\textbf{71.94}  &\textbf{72.53} \\
            \bottomrule
        \end{tabular}
        }
    \end{subtable}%
        \begin{subtable}{.33\linewidth}
      \centering
        \caption{DTD.}
        \resizebox{!}{1.3cm}{
        \begin{tabular}{l|cc|c}
            \toprule
             & Base & New & H \\
            \midrule
            CLIP & 53.12 & 55.92  & 54.48\\
            CoOp & 66.71 & 41.70 & 51.32\\
            CoCoOp & 66.67 & 41.14 &50.88 \\
            ProGrad & \underline{67.71} &50.40  &57.79 \\
            KgCoOp & 66.48 &\underline{55.28}  & \underline{60.36}\\\hline
            Ours & \textbf{68.48} &\textbf{56.81}  &\textbf{62.1} \\
            \bottomrule
        \end{tabular}
        }
    \end{subtable}%
    \begin{subtable}{.33\linewidth}
      \centering
        \caption{UCF101.}
        \resizebox{!}{1.3cm}{
        \begin{tabular}{l|cc|c}
            \toprule
             & Base & New & H \\
            \midrule
            CLIP & 68.51 & \textbf{69.77}  & 69.13\\
            CoOp & 73.30 & 57.62 & 64.52\\
            CoCoOp & 73.08 & 56.52 &63.74 \\
            ProGrad & \underline{74.87} &63.17  &68.52 \\
            KgCoOp & 73.65 &67.57  & \underline{70.48}\\\hline
            Ours & \textbf{77.94} &\underline{68.32}  &\textbf{72.75} \\
            \bottomrule
        \end{tabular}
        }
    \end{subtable}%
\label{table4}
\end{table*}

\textbf{Training Details.} 
%
We adopt ResNet-50 \cite{resnet} as the backbone for the image encoder, with 16 context tokens $M$, following \cite{coop, prograd}. 
The same training epochs, schedule, and data augmentation settings as \cite{coop, prograd} are used. 
For all experiments, we set $\lambda_1 = 1$, $\lambda_2 = 2$, and the two-layer MI estimator has hidden dimensions of 256. 
To ensure fairness, we compute the final performance at least three times with different seeds and average the results. 
All experiments are carried out on NVIDIA RTX 3090s. For more detailed training settings, please refer to the Appendix.

\textbf{Baselines.} 
We compare our method with 5 baselines: 
1) Zero-shot CLIP \cite{clip}, which employs the hand-crafted prompts. 
2) CoOp \cite{coop}, which replaces the hand-crafted prompts with a set of learnable prompts trained on the downstream tasks. 
3) CoCoOp \cite{cocoop}, which improves CoOp by generating the image-conditioned vectors combined with learnable prompts. 
4) ProGrad, which optimizes learnable prompts based on hand-crafted gradients. 
5) KgCoOp, which regularizes the learnable text embeddings with hand-crafted text embeddings.

\begin{table*}[t]
\caption{
Quantitative comparisons for the domain generalization, using 4-shot from source. 
}
\vspace{-1em}
\begin{tabular}{l|c|ccccc}
\hline
\multirow{2}{*}{\textbf{Method}} & \textbf{Source} & \multicolumn{5}{c}{\textbf{Target}}                                                                      \\ \cline{2-7} 
                                 & ImageNet        & ImageNetV2     & ImageNet-Sketch & ImageNet-A     & \multicolumn{1}{c|}{ImageNet-R}     & Average          \\ \hline
CLIP                             & 58.18           & 51.34          & 33.32           & 21.65          & \multicolumn{1}{c|}{56.00}          & 44.10           \\
CoOp                             & 59.51           & 51.74          & 30.99           & 21.23          & \multicolumn{1}{c|}{53.52}          & 43.40          \\
CoCoOp                           & 61.01           & 53.87          & 32.65           & 22.56          & \multicolumn{1}{c|}{54.31}          & 44.88          \\
ProGrad                          & {\ul 61.46}     & \textbf{54.39} & 32.86           & {\ul 22.33}    & \multicolumn{1}{c|}{55.16}          & {\ul 45.44}    \\
KgCoOp                           & 59.84           & 54.02          & {\ul 33.28}     & 22.15          & \multicolumn{1}{c|}{{\ul 56.24}}    & 45.11          \\ \hline
Ours                             & \textbf{61.53}  & {\ul 54.36}    & \textbf{35.14}  & \textbf{23.01} & \multicolumn{1}{c|}{\textbf{58.28}} & \textbf{46.46} \\ \hline
\end{tabular}
\label{table5}
\vspace{-1em}
\end{table*}

\subsection{Generalization From Base to New Classes} \label{subsec:base-new}
As in \cite{coop, cocoop}, we partition each dataset into two non-overlapping sets of classes: the ``base'' classes and the ``new'' classes.
%
To evaluate the generalization ability, all methods are trained on the base classes with a few shot samples for prompt tuning and evaluated on the base and new classes. 
The class-wise accuracy and Harmonic mean metrics (H, i.e., $\rm{HM} = \frac{{2 \times {A_b} \times {A_n}}}{{{A_b} + {A_n}}}$) are employed. We conduct experiments on the 1, 2, 4, 8, and 16 shots for comprehensive evaluations. 
The average performance among all 11 datasets is illustrated in Table~\ref{table3} and the detailed performance on all 11 datasets with 4 shots is illustrated in Table~\ref{table4}.

Table~\ref{table3} shows that our method outperforms existing methods in terms of Harmonic mean across all few-shot settings, demonstrating a better trade-off between task-specific and general abilities.
%
Our method also achieves the highest performance on new classes across all settings and the highest performance on base classes for 1, 2, and 4 shot settings.
%
While CoOp-based methods fail to preserve general ability compared to ProGrad and KgCoOp, KgCoOp performs worse than CoOp on base classes with 8 and 16 shots.
%
One possible reason is that the adaptability of the learnable text prompt to the training samples is hindered by the Euclidean distance between the hand-crafted text embeddings and the learnable text embeddings, as shown in Equation (5).
%
In terms of the trade-off between task-specific and general abilities, across all shots settings, KgCoOp surpasses ProGrad on new classes and fails on base classes. 
%
For instance, compared with ProGrad, KgCoOp improves the accuracy on new classes by 3.12$\%$ while dropping on base classes by 2.41$\%$ for the 16-shot setting. 
%
ProGrad slightly exceeds ours on base classes for the 8 and 16-shot settings while obtaining a worse performance on new classes, indicating the generated prompts are serious overfitting on base classes. 
In summary, our proposed approach demonstrates superior performance across 11 datasets, which confirms that it strikes an optimal balance between adapting task-specific ability to downstream tasks and retaining the general ability of pre-trained VLMs for unseen classes.

Table~\ref{table4} presents a detailed analysis of the performance of existing methods on 11 datasets.
%
All methods exhibit a significant improvement in the accuracy on base classes compared to zero-shot CLIP.
%
Specifically, CoOp, CoCoOp, ProGrad, and KgCoOp surpass CLIP by 6.84$\%$, 6.68$\%$, 8.86$\%$, and 7.29$\%$, respectively.
%
However, the improved task-specific ability of previous methods results in a bias towards base classes, leading to a significant drop in the general (zero-shot) ability on new classes.
%
As results, previous methods exhibit a drop of 11.40$\%$, 8.93$\%$, 5.30$\%$, and 1.02$\%$, respectively.
%
Our proposed method outperforms prompt tuning methods on new classes as well as zero-shot CLIP, thanks to the class-wise augmentation.
%
We achieve the best accuracy on new classes for 6 out of 11 datasets and the best accuracy on base classes for 8 out of 11 datasets.

\subsection{Domain Generalization} \label{subsec:domain}
Generalizing to out-of-distribution (OOD) data is a crucial skill for machine learning algorithms \cite{ood}.
%
As with CoCoOp and ProGrad, we perform prompt tuning on the source dataset (ImageNet \cite{imagenet}) with a few shots and then evaluate the model on the target datasets (ImageNetV2 \cite{v2}, ImageNet-Sketch \cite{sketch}, ImageNet-A \cite{a}, and ImageNet-R \cite{r}).
%
These five datasets have identical classes, but the target datasets have different data distributions from the source dataset.

Table \ref{table5}~illustrates the results. 
%
Our method outperforms all other methods, achieving the best performance on 4 out of 5 datasets, and obtaining the best average accuracy.
%
CoOp and CoCoOp fail in both source and target datasets, indicating that these methods may not fully adapt to downstream tasks and undermine general ability.
%
KgCoOp performs well on OOD datasets but fails on the source dataset, similar to the trend in Section~\ref{subsec:base-new}.
%
ProGrad demonstrates greater domain generalizability than CoOp, CoCoOp, and KgCoOp. 
%
Compared to ProGrad, our proposed method achieves a more balanced performance on both source and target datasets.

\begin{table*}[t]
\caption{
Quantitative comparisons for the cross-dataset transferability, using 4-shot from the source.
} 
\vspace{-1em}
\begin{tabular}{l|c|ccccccccccc}
\hline
\multirow{2}{*}{\textbf{Method}} & \textbf{Source} & \multicolumn{11}{c}{\textbf{Target}}                                                                                                                                                                          \\ \cline{2-13} 
                                 & \rotatebox{90}{ImageNet}        & \rotatebox{90}{Caltech101}     & \rotatebox{90}{OxfordPets}     & \rotatebox{90}{StandfordCars}  & \rotatebox{90}{Flowers102}     & \rotatebox{90}{Food101}        & \rotatebox{90}{FGVCAiraft}     & \rotatebox{90}{EuroSAT}        & \rotatebox{90}{SUN397}         & \rotatebox{90}{DTD}            & \multicolumn{1}{c|}{\rotatebox{90}{UCF101}}         & \rotatebox{90}{Average}        \\ \hline
CoOp                             & 61.01           & 81.58          & 82.77          & 48.19          & 58.91          & 72.33          & 9.78           & {\ul 28.93}    & 55.16          & 30.73          & \multicolumn{1}{c|}{51.56}          & 52.81          \\
CoCoOp                           & 60.43           & 82.78          & 83.36          & 48.86          & 62.59        & 74.25          & 13.82          & 27.84          & 57.86          & 35.82          & \multicolumn{1}{c|}{53.36}          & 54.66          \\
ProGrad                          & {\ul 61.46}     & 86.17          & \textbf{85.50} & {\ul 54.26}    & 62.77          & 74.28          & 14.13          & 22.14          & 56.07          & 36.47          & \multicolumn{1}{c|}{55.25}          & 55.31          \\
KgCoOp                           & 59.84           & {\ul 87.55}    & {\ul 84.93}    & \textbf{55.08} & \textbf{64.15} & \textbf{76.57} & {\ul 14.36}    & 25.99          & {\ul 60.28}    & {\ul 36.80}    & \multicolumn{1}{c|}{{\ul 58.22}}    & {\ul 56.69}    \\
Ours                             & \textbf{61.53}  & \textbf{89.01} & 83.81          & 53.79          & {\ul62.87}          & {\ul 75.80}    & \textbf{15.66} & \textbf{30.22} & \textbf{60.77} & \textbf{39.48} & \multicolumn{1}{c|}{\textbf{59.21}} & \textbf{57.47} \\ \hline
\end{tabular}
\label{table6}
\vspace{-1em}
\end{table*}

\subsection{Cross-dataset Transferability} \label{subsec:cross}
We have shown that our approach is generalizable within a dataset.
However, transferring to a different dataset can be more challenging since the underlying principles can change completely, such as from object recognition to texture classification.
%
To address this, we follow \cite{cocoop} and tune the text prompt on the source dataset (ImageNet) and evaluate on 10 target datasets.
%
Table~\ref{table6} shows the performance of compared prompt tuning methods.
%
Our proposed method achieves the best results in 7 out of 11 datasets and has the highest average accuracy.
%
CoOp and CoCoOp are lacking in general knowledge as they are biased to seen classes. 
%
ProGrad adapts to the source dataset but does not perform well in the target datasets.
%
KgCoOp has the excellent general ability and achieves the best results in StandfordCars, Flowers102, and Food 101 datasets while failing in the source dataset, which is consistent with the previous experiments.

\begin{table}[] \small
\caption{
Ablation studies.
%
``Baseline'' refers to the model that only includes the learnable prompt (degrades to CoOp).
``MI loss'' denotes utilizing Equation (13) to optimize the network. 
``Aug'' denotes class-wise augmentation strategy.
} 
\vspace{-1em}
\begin{tabular}{c|ccc|ccc}
\hline
\multirow{2}{*}{Method} & \multicolumn{3}{c|}{K=4}                         & \multicolumn{3}{c}{K=16}                         \\ \cline{2-7} 
                        & Base           & New            & H              & Base           & Novel          & H              \\ \hline
Baseline                & 72.06          & 59.69          & 65.29          & 77.24          & 57.40          & 65.86          \\ \hline
ProGrad                & 73.88          & \cellcolor{lightgray}64.95          & 69.13          & 77.98          & \cellcolor{lightgray}64.41          & 69.94          \\ 
KgCoOp                & \cellcolor{lightgray}72.42         & 68.00          & 70.14          & \cellcolor{lightgray}75.51          & 67.53          & 71.30          \\ \hline
+MI loss                & 74.23          & 68.23          & 71.08          &  77.21         & 67.43          & 71.56               \\ 
+MI loss+Aug            & 74.77          & 69.14          & 71.65          & 77.60          & 68.36          & 72.12           \\ \hline
\end{tabular}
\label{table7}
\vspace{-1em}
\end{table}

\begin{table}[]
\caption{Training efficiency comparisons. The \emph{time} indicates the average time to train one image (i.e., ms/image).} 
\vspace{-1em}
\begin{tabular}{c|ccccc}
\hline
\multicolumn{1}{l|}{} & CoOp  & CoCoOp & ProGrad & KgCoOp & Ours  \\ \hline
\emph{time}         & 1.8ms & 47ms  & 6.5ms    & 1.8ms  & 4.5ms  \\ 
H                   & 65.29 & 68.45 & 69.13   & 70.14  & 71.65 \\ \hline
\end{tabular}
\label{table8}
\vspace{-1em}
\end{table}

\section{Ablation Study}

\subsection{Effect of Different Components}
\begin{figure}[t]
\centering
\includegraphics[width=0.5\textwidth]{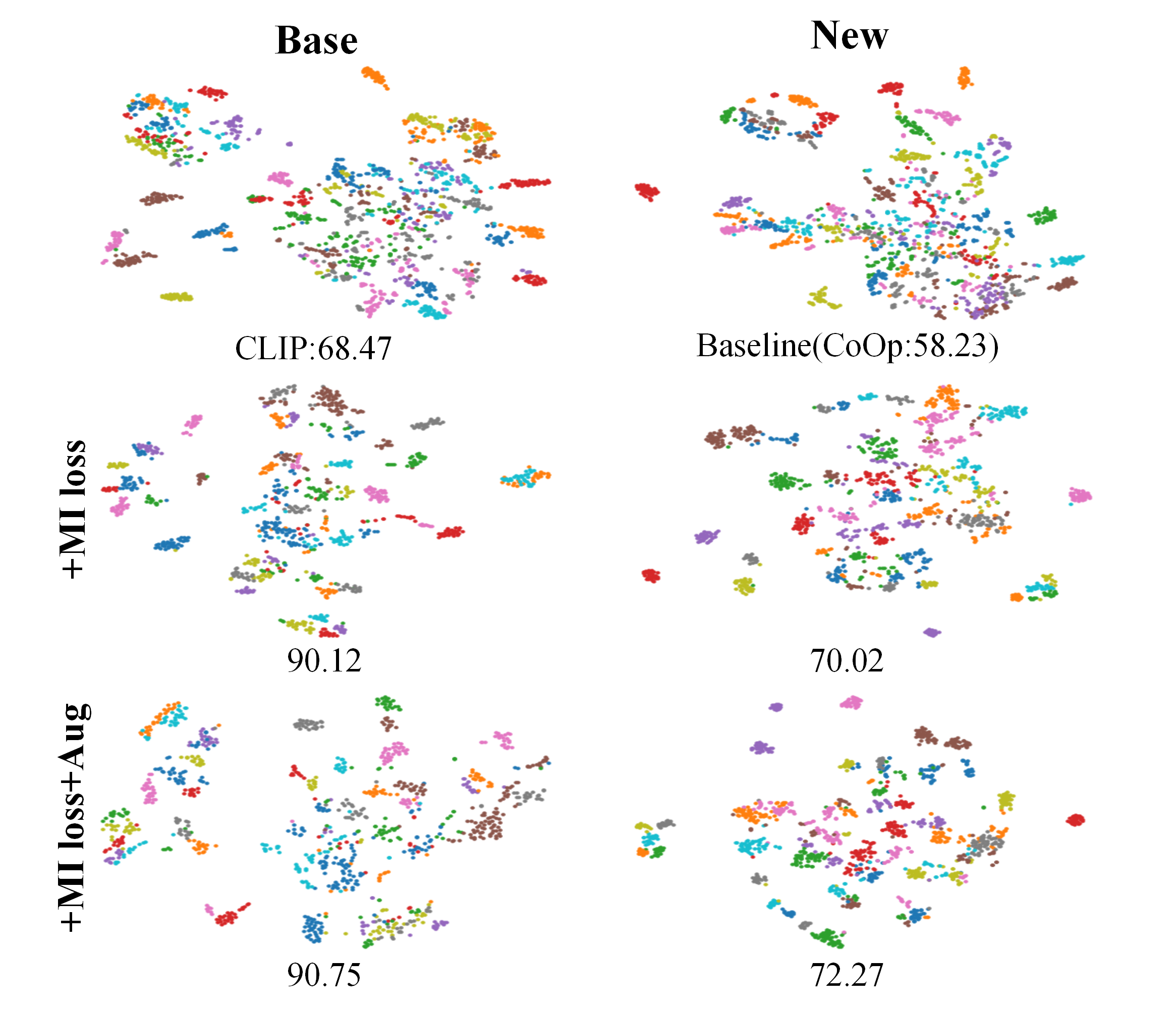}
\caption{t-SNE visualization on Flowers102 \cite{flowers}.}
\label{fig2}
\vspace{-1em}
\end{figure}

We evaluate the effectiveness of each component of the proposed method.
%
The results are shown in Table~\ref{table7}.
Both ProGrad and KgCoOp introduce general knowledge from hand-crafted prompts and significantly improve performance, particularly in new classes. 
%
However, these methods are unable to achieve a desirable trade-off between the base classes and new classes (marked in gray).
%
KgCoOp employs a relatively harder distance constraint (i.e., Euclidean distance) compared to ProGrad, which effectively improves generalization ability but hinders the model's ability to adapt to specific information.
%
ProGrad modulates the gradient directions but cannot fully utilize general information.
%
Applying MI to ensemble the task-specific and general information achieves a better trade-off between the base classes and new classes by a clear margin.
%
Moreover, the class-wise augmentation from the visual modality enhances the expressiveness of the prompt subspace, resulting in significant robustness to a broader range of new classes. 
%
Figure~\ref{fig2} presents t-SNE visualizations of proposed MI loss and the class-wise augmentation strategy, showing that MI loss significantly improves inter-class boundaries and intra-class clustering compared with zero-shot CLIP and CoOp.

\subsection{Training Efficiency} \label{subsec:te}
Table~\ref{table8} presents the training time of prompt tuning methods with 4 shots, where the training time is calculated per image (i.e., $ms$/image).
%
Our proposed method has a significantly lower training time compared to CoCoOp (4.5$ms$ vs 47$ms$), and it is also faster than ProGrad.
%
CoOp and KgCoOp only optimize the learnable prompt, which takes the least amount of time.
%
It is worth noting that our method does not introduce additional parameters during the inference phase, unlike CoCoOp. 
The learnable MI estimator in our method is only used during the training phase to facilitate MI estimation and is ablated during the inference phase.

\section{Conclusion and Future Work}
In this work, we presented a novel framework as a way of learning generalizable prompts for vision-language models (VLMs) like CLIP, retaining the generalization ability without losing the downstream task performance benefits.
We leveraged the hand-crafted prompts and soft prompts as dual views to ensemble semantic information from the textual modality, without introducing extra network modules or laborious operations in the inference phase.
Moreover, given that previous studies have mainly concentrated on the textual modality, we utilized cross-class information to tap into the potential for generalization from the visual modality, an area that has not been extensively explored in previous research.
On the other hand, enriching the prompt subspace for VLMs in an unsupervised fashion is still an open area that will help with the usability and expressivity of prompt tuning.
Class-wise augmentation provides an effective interface, yet it is unclear what is the optimal way.
Here, we presented a simple way of taking enriched prompt subspace via multimodal ensemble and exploration; how to incorporate richer prompt subspace is an interesting direction for future work.

\bibliographystyle{ACM-Reference-Format}
\bibliography{sample.bib}

\end{document}